\pdfoutput=1

\documentclass[11pt]{article}

\usepackage[final]{acl}

\usepackage{times}
\usepackage{latexsym}

\usepackage[T1]{fontenc}

\usepackage[utf8]{inputenc}

\usepackage{microtype}

\usepackage{inconsolata}

\usepackage{graphicx}

\usepackage{comment}

\title{Does Training on Synthetic Data Make Models Less Robust?}

\author{Lingze Zhang \and Ellie Pavlick\\
  Brown University \\
  \texttt{\{lingze\_zhang, ellie\_pavlick\}@brown.edu} \\}

\begin{document}
\maketitle

\begin{abstract}
An increasingly common practice is to train large language models (LLMs) using synthetic data. Often this synthetic data is produced by the same or similar LLMs as those it is being used to train. This raises the question of whether the synthetic data might in fact exacerbate certain ``blindspots'' by reinforcing heuristics that the LLM already encodes. In this paper, we conduct simulated experiments on the natural language inference (NLI) task with Llama-2-7B-hf models. We use MultiNLI as the general task and HANS, a targeted evaluation set designed to measure the presence of specific heuristic strategies for NLI, as our ``blindspot'' task. Our goal is to determine whether performance disparities between the general and blind spot tasks emerge. Our results indicate that synthetic data does not reinforce blindspots in the way we expected. Specifically, we see that, while fine-tuning with synthetic data doesn't necessarily reduce the use of the heuristic, it also does not make it worse as we hypothesized. \footnote{Our code is available at \url{https://github.com/untakenJ/synthetic-data-blindspot}.}
\end{abstract}

\section{Introduction and Related Work}
Constructing a dataset for a specific task in natural language processing can be costly in terms of time and labor. An increasingly common approach to solve this problem is to take advantage of large language models (LLMs) to generate training data. It’s simple to fine-tune an LLM or just use in-context learning to generate huge amounts of training data with a relatively small number of demonstrations. However, how effective the model-written datasets are for different tasks is still an open question.

Model-generated training data is widely used in different domains like image classification~\cite{9053146,gowal2021improving}, visual language concepts understanding~\cite{cascante2023going} and medical image understanding~\cite{fernandez2022can}. In many NLP tasks, such as commonsense reasoning~\cite{yang-etal-2020-generative}, question-answering~\cite{bartolo-etal-2021-improving,paranjape-etal-2022-retrieval}, sycophancy reduction~\cite{wei2024simplesyntheticdatareduces}, cultural debiasing~\cite{li2024culturellmincorporatingculturaldifferences,li2024cultureparkboostingcrossculturalunderstanding} and general instruction alignment~\cite{wang-etal-2023-self-instruct}, synthetic data created with generative models are utilized in model training. In cases where there are limited sources for model training, synthetic data would greatly benefit the performance of finetuned model. High-quality model-written datasets may also be used for evaluations. \citet{perez-etal-2023-discovering} created 154 evaluation datasets and discovered inverse scaling of language models in some scenarios.

However, synthetic data may also be harmful. \citet{shumailov2024ai} found that language models may collapse if recursively finetuned with generated text. Such degradation has also been discovered in image generation tasks~\cite{alemohammad2024selfconsuming}. The use of synthetic data is also criticized from the perspective of ethics and social impact~\cite{Susser2024-SUSCPF}. There's a series of research about what bias is manifested in synthetic data and how the performance in specific tasks is affected. For example, gender stereotype is a common kind of bias amplified in data generated by language models~\cite{kirk2021bias,kotek2023gender}. \citet{li-etal-2023-synthetic} investigated the text classification task and showed that subjectivity is a matter affecting the performance of models trained with synthetic data.~\citet{bisbee2023synthetic} found less variation in ChatGPT responses than in the real ANES survey. Similarly, a study from~\citet{chen-etal-2024-unveiling-flaws} indicates that the uniform format of synthetic data can lead to pattern overfitting and thus harm the instruction-following capabilities of the model trained with it.~\citet{seddik2024how} reveals that the recursive training loop makes the tails of the original distribution disappear and makes the model forget the real distribution from a statistical perspective.


One particular way in which synthetic data might be harmful is if it reinforces ungeneralizable heuristics. It is well know that LLMs often rely on features that perform well on the training set but do not necessarily generalize as we would like, for example, relying on gender bias~\cite{10.1145/3582269.3615599}, word-overlap bias in NLI~\cite{rajaee-etal-2022-looking}, or exhibiting a preference toward longer responses in text generation~\cite{singhal2024a}. We refer to these types of heuristics as \textit{blindspots}.  

In this work, we hypothesize that, because synthetic data less diverse than the original training data \cite{whitney2024real}, it is more likely to have blindspots and thus that fine-tuning on model-generated data will exacerbate these blindspots in the tuned model. In particular, we hypothesize that the synthetic data will encode the heuristic to a larger extent than naturally occurring data would, and thus that fine-tuning on synthetic data will lead the model to more strongly favor the heuristic. This weakness would be revealed in data that specifically is designed to test whether models are using the heuristic, as models trained on synthetic data might still show improved performance on generic test sets on which the heuristic performs well. 

As a case study, we focus on the natural language inference (NLI) task evaluated with the MultiNLI dataset~\cite{williams-etal-2018-broad}. The MultiNLI dataset covers general examples collected from various sources, but models trained on MultiNLI may tend to make judgments based on superficial syntactic properties and perform badly on HANS, an adversarial dataset created with syntactic heuristics~\cite{mccoy-etal-2019-right}. The HANS task can be regarded as a measure of the model's ``blindspot''.

Our expected result is that finetuning an NLI model with synthetic MultiNLI-like data will reduce its performance on HANS while improving its performance on the MultiNLI test set. However, we observed that this is not a consistent pattern under various settings of starting point model and size of synthetic dataset, though some biases do exist in the synthetic dataset. Our hypothesis is thus not fully supported by the experimental results. We have nonetheless discovered different patterns of performance change on both test sets in different scenarios. We hope the discovered insights will foster novel research ideas in understanding model degradation with synthetic training data and advancing fairness and robustness of language models.

\section{Methods}
\subsection{Overview}

We assume a \textbf{t}ask model \(T\) and a \textbf{g}enerator model \(G\).~\(T\) can be a model for any kind of NLP tasks, and \(G\) is a language model used to generate training examples for \(T\). Let $\mathcal{X}_T$ denote the set of all possible input of model $T$. The existence of a blindspot means that there's a non-random subset $\widetilde{\mathcal{X}_T}\subseteq\mathcal{X}_T$  on which the model \(T\) performs worse than on $\mathcal{X}_T$ in general. 
Let $\mathcal{D}^G$ denote the synthetic dataset generated by \(G\), and $T^{\mathcal{D}^G}$ denote the model fine-tuned on $\mathcal{D}^G$.
Our hypothesis is that $T^{\mathcal{D}^G}$ will perform worse than $T$ on $\widetilde{\mathcal{X}_T}$, but better than $T$ on $\mathcal{X}_T$.

\subsection{Tasks, Models and Datasets} \label{tmd}
\subsubsection{Tasks}
In this study, we focus on the natural language inference (NLI) task. An input example of this task contains a premise sentence, a hypothesis sentence, and a label indicating the relationship between the two sentences. The label can be one of $\{entailment, neutral, contradictory\}$.

\subsubsection{Models and Input}
Our experiments are based on the Llama-2-7B-hf model~\cite{touvron2023llama}. We fine-tuned a Llama-2 model with a classification head on top with MultiNLI as the task model $T$. The input sequence is constructed with the template 

\begin{quote}
    \textit{Please indicate the relationship between the premise and the hypothesis with entailment, neutral or contradiction. Premise: <premise> Hypothesis: <hypothesis> The relationship between premise and hypothesis is}
\end{quote} and the classification is based on the embedding of the last token in the input sequence. Our generator $G$ is also a Llama-2-7B-hf model fine-tuned with MultiNLI. It's tuned to generate examples in the form of 
\begin{quote}
\textit{This is an example where the relationship between the premise and the hypothesis is <label>. Premise: <premise> Hypothesis: <hypothesis> -- This is the end of the example.} 
\end{quote}
The label is put before the premise and the hypothesis for more flexible control of generated labels.

\subsubsection{Datasets}
We use MultiNLI \cite{williams-etal-2018-broad} as a measure of the models performance on NLI in general. The original task model $T$ and generator $G$ are both Llama-2-7B-hf models finetuned on MultiNLI. To measure the presence of the ``blindspot'', we use HANS~\cite{mccoy-etal-2019-right}. HANS is an NLI dataset created adversarially with three heuristics:  the lexical overlap heuristic, the subsequence heuristic, and the constituent heuristic. Poor performance on HANS indicates that the model is likely using these heuristics to solve the NLI task. 

When training the task model $T$, we used the training set of MultiNLI as the training data, with 750 examples (250 for each label) excluded as the dev set. HANS is not used in training at all, but the results on its test set are reported. The maximum training set size for $T$ is 391,722.

Note that there are only two labels in HANS (because of how the dataset is constructed): \textit{entailment} and \textit{non-entailment}. In our experiments, the base task model and generator are fine-tuned with three labels of MultiNLI. When testing on HANS, predicted labels \textit{neutral} and \textit{contradictory} are both regarded as \textit{non-entailment}.

\subsection{Experiments}
\subsubsection{Basic Setting}
In our experiment pipeline, we first fine-tuned a classifier model $T$ with the MultiNLI training set from the pretrained Llama-2-7B-hf model with a classification head. Then we fine-tuned another Llama-2-7B-hf model as the generator $G$, also with the MultiNLI training set. After training $G$, we generated a dataset $\mathcal{D}^G$ with it and used $\mathcal{D}^G$ to further fine-tune $T$ to obtain the further tuned model $T^{\mathcal{D}^G}$. We varied $T$ (by changing the number of MultiNLI examples used for the initial fine-tuning) and $\mathcal{D}^G$ for different settings.

\subsubsection{Starting Models}
The initial task model $T$ is fine-tuned with data from the original MultiNLI dataset. In order to simulate task models in different stages, we trained 6 starting models with training set sizes of 0 (meaning the official pretrained model with a random classification head), 5000, 10000, 20000, 100000, and 391722.

\subsubsection{Synthetic Datasets}
The synthetic data examples are all generated by a Llama-2-7B-hf model $G$ fine-tuned with the MultiNLI training set for 1 epoch. The generator model is fine-tuned to generate text in the specific format aforementioned with the following prompt:

\begin{quote}
    \textit{This is an example where the relationship between the premise and the hypothesis is <label>}
\end{quote} We kept the generated examples in which the premise and the hypothesis can be extracted with a regular expression without further filtering. 

We generated 1,819,813 examples, which is more than necessary for the training. We sampled two kinds of synthetic datasets: uniformly random sampled datasets and showcasing datasets with a stronger bias. We took the lexical overlap (LO) heuristic addressed in the HANS dataset as an example. Lexical overlap means all the words in the hypothesis appear in the premise. 

Based on the availability of synthetic data, we constructed synthetic training sets of three sizes: 73080, 36040, and 18020. In each synthetic set, there are equal numbers of examples with each label. The random synthetic dataset (marked as \textit{Synthetic}) is uniformly sampled for each label, and the more strongly biased dataset (\textit{Biased Synthetic}) is sampled to make sure all entailment examples follow the lexical overlap heuristic and all other examples do not. We also included baseline datasets sampled from the original MultiNLI training set of the same sizes, marked as \textit{Original}. The datasets used in the experiment can be represented as $\{73080, 36040, 18020\} \times \{Original, Synthetic, Biased\ Synthetic\}$.

\begin{figure*}[t]
  \centering
  \includegraphics[width=1.0\linewidth]{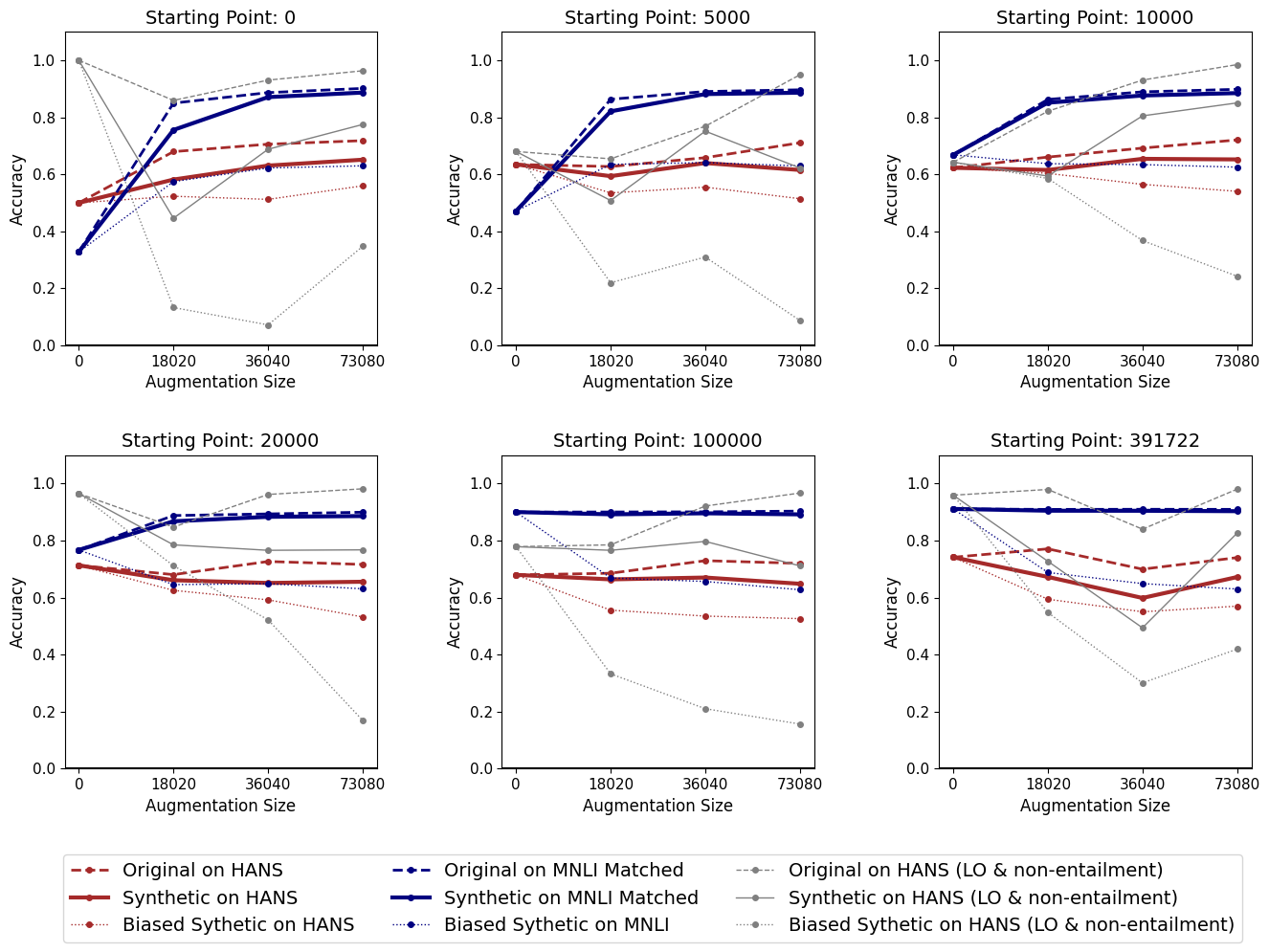} 
  \caption {Augmented model performance under different settings.}
  \label{fig:results}
\end{figure*}

\subsubsection{Test Sets}
We report our results on three test sets: the MultiNLI Matched test set, the HANS test set, and the subset of the HANS test set with lexical overlap and a non-entailment label, which reflects the model's performance specifically on the blind spot. In addition to the augmented model with different training sets, we also report the classification performance of each starting model.

\section{Results}
Our main results are reported in Figure \ref{fig:results}. Each subplot corresponds to a different starting model $T$. 
When starting with undertrained task models, further fine-tuning with synthetic data will improve the performance on the MultiNLI Matched test set. The amount of improvement is on par with the model fine-tuned with original MultiNLI training data if the training set size is large enough. For relatively well-established starting models, neither fine-tuning with original nor synthetic data would significantly improve the performance of MultiNLI.

The performance on the HANS test set is trickier. The hypothesized trend , in which the performance of HANS goes down while the performance of MultiNLI goes up, only happens in for the 20K starting point. We also see a fairly sizable drop in HANS performance for the 392K starting point, but the curve is not monotonic and thus it is inconclusive. Overall, under most settings, further fine-tuning with original MultiNLI data would always benefit more or harm less on HANS performance than synthetic data. The gap does exist, but may not be as serious as expected.

As a sanity check, we also trained the model with the biased synthetic dataset in which all examples with lexical overlap are labeled entailment, and no example with neural or contradiction label satisfies the lexical overlap heuristic. As expected, such models perform worst in almost all tests, with the accuracy on the HANS subset of lexical overlap heuristic and non-entailment label dropping significantly towards zero over training. This indicates that a very biased synthetic dataset could exacerbate blindspots as expected, and thus implies that true synthetic data does not overrepresent the heuristic as much as hypothesized.


\section{Conclusion}
From the simulated experiments, we observed that while training the task model with synthetic data contributes to the performance on the general tasks almost equally compared with training with the original data, the contribution gap on the ``more difficult'' blindspot task does exist. Under some settings, there's a dispersion where the accuracy on the blindspot task goes down while the general task accuracy goes up, but this is not a consistent tendency. Reinforcement of bias while training may happen, but this would probably not cause significant issues if we just use the unfiltered synthetic data for training. 

\newpage

\section*{Limitations}

\begin{table}[h]
\small
\centering
\begin{tabular}{rrrrrr}
\hline
\textbf{Dataset} & \textbf{Label} & \textbf{Count} & \textbf{\# LO} & \textbf{\% LO} & \textbf{\% S:O} \\ \hline
                 & Ent.           & 631992         & 24360          & 3.854          & 2.352           \\
Syn.             & Neu.           & 562822         & 690            & 0.123         & 2.320           \\
                 & Con.           & 624999         & 1504           & 0.241        & 1.827           \\ \hline
                 & Ent.           & 130541         & 2139           & 1.639         & -               \\
Orig.            & Neu.           & 130573         & 69             & 0.053        & -               \\
                 & Con.           & 130608         & 172            & 0.132         & -               \\ \hline
\end{tabular}
\caption{\label{tab:dstat}
Statistics of examples with lexical overlap in original and synthetic data. There are generally more cases in synthetic data, and the correlation between lexical overlap and the entailment label is reinforced.
  }
\end{table}

We need to note that the study with MultiNLI and HANS is a case study addressing the issue of bias reinforcement when training models with synthetic data. It's still an open question whether the results about the biases we are studying are generalizable to other cases. According to Table \ref{tab:dstat}, lexical overlap is more common in the synthetic dataset than in MulitNLI for all labels, which may indicate that synthetic data is less diverse. However, the correlation between lexical overlap and entailment label is just slightly stronger. Different kinds of bias can emerge in very different ways in synthetic data, and this makes it challenging to evaluate the effect of training models with synthetic data holistically.

Our design choices about the experiments may also be arbitrary. The task model we choose is the Llama-2-7B-hf model with a classification head. The pretrained Llama model is a relatively strong model, while the initialization of the classification head is random. Whether jointly training these parts is a reasonable choice is still arguable. Moreover, the two-step approach of model training is also not the only choice. It's also common to mix the original and synthetic data in different ratios and train the model with the mixed dataset in one run. Varying the experiment design is also necessary for further validations about the findings in this study.

Another notable point is that pretrained large language models, such as Llama, encode a wealth of world knowledge. Many potential biases may have been mitigated during training, particularly for models deployed in real-world applications, which are typically much larger and more powerful than the fine-tuned Llama-2-7B-hf generator used in our study. On the other hand, human-created or audited data are not inherently free from implicit biases. The more concentrated distribution and reduced diversity of synthetic data might reinforce biases in certain blindspot scenarios. However, the rich world knowledge embedded in the generator model can also address some biases, potentially outperforming humans in certain cases and contributing positively to bias mitigation. A critical direction for future research is to disentangle these two effects and assess the significance of each, thereby enhancing our understanding of the impact of training models with synthetic data.

\bibliography{anthology,custom}

\begin{thebibliography}{28}
\providecommand{\natexlab}[1]{#1}

\bibitem[{Alemohammad et~al.(2024)Alemohammad, Casco-Rodriguez, Luzi, Humayun, Babaei, LeJeune, Siahkoohi, and Baraniuk}]{alemohammad2024selfconsuming}
Sina Alemohammad, Josue Casco-Rodriguez, Lorenzo Luzi, Ahmed~Imtiaz Humayun, Hossein Babaei, Daniel LeJeune, Ali Siahkoohi, and Richard Baraniuk. 2024.
\newblock \href {https://openreview.net/forum?id=ShjMHfmPs0} {Self-consuming generative models go {MAD}}.
\newblock In \emph{The Twelfth International Conference on Learning Representations}.

\bibitem[{Bartolo et~al.(2021)Bartolo, Thrush, Jia, Riedel, Stenetorp, and Kiela}]{bartolo-etal-2021-improving}
Max Bartolo, Tristan Thrush, Robin Jia, Sebastian Riedel, Pontus Stenetorp, and Douwe Kiela. 2021.
\newblock \href {https://doi.org/10.18653/v1/2021.emnlp-main.696} {Improving question answering model robustness with synthetic adversarial data generation}.
\newblock In \emph{Proceedings of the 2021 Conference on Empirical Methods in Natural Language Processing}, pages 8830--8848, Online and Punta Cana, Dominican Republic. Association for Computational Linguistics.

\bibitem[{Besnier et~al.(2020)Besnier, Jain, Bursuc, Cord, and Pérez}]{9053146}
Victor Besnier, Himalaya Jain, Andrei Bursuc, Matthieu Cord, and Patrick Pérez. 2020.
\newblock \href {https://doi.org/10.1109/ICASSP40776.2020.9053146} {This dataset does not exist: Training models from generated images}.
\newblock In \emph{ICASSP 2020 - 2020 IEEE International Conference on Acoustics, Speech and Signal Processing (ICASSP)}, pages 1--5.

\bibitem[{Bisbee et~al.(2024)Bisbee, Clinton, Dorff, Kenkel, and Larson}]{bisbee2023synthetic}
James Bisbee, Joshua~D Clinton, Cassy Dorff, Brenton Kenkel, and Jennifer~M Larson. 2024.
\newblock Synthetic replacements for human survey data? the perils of large language models.
\newblock \emph{Political Analysis}, pages 401--416.

\bibitem[{Cascante-Bonilla et~al.(2023)Cascante-Bonilla, Shehada, Smith, Doveh, Kim, Panda, Varol, Oliva, Ordonez, Feris et~al.}]{cascante2023going}
Paola Cascante-Bonilla, Khaled Shehada, James~Seale Smith, Sivan Doveh, Donghyun Kim, Rameswar Panda, Gul Varol, Aude Oliva, Vicente Ordonez, Rogerio Feris, et~al. 2023.
\newblock Going beyond nouns with vision \& language models using synthetic data.
\newblock In \emph{Proceedings of the IEEE/CVF International Conference on Computer Vision}, pages 20155--20165.

\bibitem[{Chen et~al.(2024)Chen, Zhang, Wang, Zhao, Wen, and Chen}]{chen-etal-2024-unveiling-flaws}
Jie Chen, Yupeng Zhang, Bingning Wang, Xin Zhao, Ji-Rong Wen, and Weipeng Chen. 2024.
\newblock \href {https://doi.org/10.18653/v1/2024.findings-emnlp.873} {Unveiling the flaws: Exploring imperfections in synthetic data and mitigation strategies for large language models}.
\newblock In \emph{Findings of the Association for Computational Linguistics: EMNLP 2024}, pages 14855--14865, Miami, Florida, USA. Association for Computational Linguistics.

\bibitem[{Fernandez et~al.(2022)Fernandez, Pinaya, Borges, Tudosiu, Graham, Vercauteren, and Cardoso}]{fernandez2022can}
Virginia Fernandez, Walter Hugo~Lopez Pinaya, Pedro Borges, Petru-Daniel Tudosiu, Mark~S Graham, Tom Vercauteren, and M~Jorge Cardoso. 2022.
\newblock Can segmentation models be trained with fully synthetically generated data?
\newblock In \emph{International Workshop on Simulation and Synthesis in Medical Imaging}, pages 79--90. Springer.

\bibitem[{Gowal et~al.(2021)Gowal, Rebuffi, Wiles, Stimberg, Calian, and Mann}]{gowal2021improving}
Sven Gowal, Sylvestre-Alvise Rebuffi, Olivia Wiles, Florian Stimberg, Dan~Andrei Calian, and Timothy~A Mann. 2021.
\newblock Improving robustness using generated data.
\newblock \emph{Advances in Neural Information Processing Systems}, 34:4218--4233.

\bibitem[{Kirk et~al.(2021)Kirk, Jun, Volpin, Iqbal, Benussi, Dreyer, Shtedritski, and Asano}]{kirk2021bias}
Hannah~Rose Kirk, Yennie Jun, Filippo Volpin, Haider Iqbal, Elias Benussi, Frederic Dreyer, Aleksandar Shtedritski, and Yuki Asano. 2021.
\newblock Bias out-of-the-box: An empirical analysis of intersectional occupational biases in popular generative language models.
\newblock \emph{Advances in neural information processing systems}, 34:2611--2624.

\bibitem[{Kotek et~al.(2023{\natexlab{a}})Kotek, Dockum, and Sun}]{kotek2023gender}
Hadas Kotek, Rikker Dockum, and David Sun. 2023{\natexlab{a}}.
\newblock Gender bias and stereotypes in large language models.
\newblock In \emph{Proceedings of the ACM collective intelligence conference}, pages 12--24.

\bibitem[{Kotek et~al.(2023{\natexlab{b}})Kotek, Dockum, and Sun}]{10.1145/3582269.3615599}
Hadas Kotek, Rikker Dockum, and David Sun. 2023{\natexlab{b}}.
\newblock \href {https://doi.org/10.1145/3582269.3615599} {Gender bias and stereotypes in large language models}.
\newblock In \emph{Proceedings of The ACM Collective Intelligence Conference}, CI '23, page 12–24, New York, NY, USA. Association for Computing Machinery.

\bibitem[{Li et~al.(2024{\natexlab{a}})Li, Chen, Wang, Sitaram, and Xie}]{li2024culturellmincorporatingculturaldifferences}
Cheng Li, Mengzhou Chen, Jindong Wang, Sunayana Sitaram, and Xing Xie. 2024{\natexlab{a}}.
\newblock \href {https://arxiv.org/abs/2402.10946} {Culturellm: Incorporating cultural differences into large language models}.
\newblock \emph{Preprint}, arXiv:2402.10946.

\bibitem[{Li et~al.(2024{\natexlab{b}})Li, Teney, Yang, Wen, Xie, and Wang}]{li2024cultureparkboostingcrossculturalunderstanding}
Cheng Li, Damien Teney, Linyi Yang, Qingsong Wen, Xing Xie, and Jindong Wang. 2024{\natexlab{b}}.
\newblock \href {https://arxiv.org/abs/2405.15145} {Culturepark: Boosting cross-cultural understanding in large language models}.
\newblock \emph{Preprint}, arXiv:2405.15145.

\bibitem[{Li et~al.(2023)Li, Zhu, Lu, and Yin}]{li-etal-2023-synthetic}
Zhuoyan Li, Hangxiao Zhu, Zhuoran Lu, and Ming Yin. 2023.
\newblock \href {https://doi.org/10.18653/v1/2023.emnlp-main.647} {Synthetic data generation with large language models for text classification: Potential and limitations}.
\newblock In \emph{Proceedings of the 2023 Conference on Empirical Methods in Natural Language Processing}, pages 10443--10461, Singapore. Association for Computational Linguistics.

\bibitem[{McCoy et~al.(2019)McCoy, Pavlick, and Linzen}]{mccoy-etal-2019-right}
Tom McCoy, Ellie Pavlick, and Tal Linzen. 2019.
\newblock \href {https://doi.org/10.18653/v1/P19-1334} {Right for the wrong reasons: Diagnosing syntactic heuristics in natural language inference}.
\newblock In \emph{Proceedings of the 57th Annual Meeting of the Association for Computational Linguistics}, pages 3428--3448, Florence, Italy. Association for Computational Linguistics.

\bibitem[{Paranjape et~al.(2022)Paranjape, Lamm, and Tenney}]{paranjape-etal-2022-retrieval}
Bhargavi Paranjape, Matthew Lamm, and Ian Tenney. 2022.
\newblock \href {https://doi.org/10.18653/v1/2022.acl-long.117} {Retrieval-guided counterfactual generation for {QA}}.
\newblock In \emph{Proceedings of the 60th Annual Meeting of the Association for Computational Linguistics (Volume 1: Long Papers)}, pages 1670--1686, Dublin, Ireland. Association for Computational Linguistics.

\bibitem[{Perez et~al.(2023)Perez, Ringer, Lukosiute, Nguyen, Chen, Heiner, Pettit, Olsson, Kundu, Kadavath, Jones, Chen, Mann, Israel, Seethor, McKinnon, Olah, Yan, Amodei, Amodei, Drain, Li, Tran-Johnson, Khundadze, Kernion, Landis, Kerr, Mueller, Hyun, Landau, Ndousse, Goldberg, Lovitt, Lucas, Sellitto, Zhang, Kingsland, Elhage, Joseph, Mercado, DasSarma, Rausch, Larson, McCandlish, Johnston, Kravec, El~Showk, Lanham, Telleen-Lawton, Brown, Henighan, Hume, Bai, Hatfield-Dodds, Clark, Bowman, Askell, Grosse, Hernandez, Ganguli, Hubinger, Schiefer, and Kaplan}]{perez-etal-2023-discovering}
Ethan Perez, Sam Ringer, Kamile Lukosiute, Karina Nguyen, Edwin Chen, Scott Heiner, Craig Pettit, Catherine Olsson, Sandipan Kundu, Saurav Kadavath, Andy Jones, Anna Chen, Benjamin Mann, Brian Israel, Bryan Seethor, Cameron McKinnon, Christopher Olah, Da~Yan, Daniela Amodei, Dario Amodei, Dawn Drain, Dustin Li, Eli Tran-Johnson, Guro Khundadze, Jackson Kernion, James Landis, Jamie Kerr, Jared Mueller, Jeeyoon Hyun, Joshua Landau, Kamal Ndousse, Landon Goldberg, Liane Lovitt, Martin Lucas, Michael Sellitto, Miranda Zhang, Neerav Kingsland, Nelson Elhage, Nicholas Joseph, Noemi Mercado, Nova DasSarma, Oliver Rausch, Robin Larson, Sam McCandlish, Scott Johnston, Shauna Kravec, Sheer El~Showk, Tamera Lanham, Timothy Telleen-Lawton, Tom Brown, Tom Henighan, Tristan Hume, Yuntao Bai, Zac Hatfield-Dodds, Jack Clark, Samuel~R. Bowman, Amanda Askell, Roger Grosse, Danny Hernandez, Deep Ganguli, Evan Hubinger, Nicholas Schiefer, and Jared Kaplan. 2023.
\newblock \href {https://doi.org/10.18653/v1/2023.findings-acl.847} {Discovering language model behaviors with model-written evaluations}.
\newblock In \emph{Findings of the Association for Computational Linguistics: ACL 2023}, pages 13387--13434, Toronto, Canada. Association for Computational Linguistics.

\bibitem[{Rajaee et~al.(2022)Rajaee, Yaghoobzadeh, and Pilehvar}]{rajaee-etal-2022-looking}
Sara Rajaee, Yadollah Yaghoobzadeh, and Mohammad~Taher Pilehvar. 2022.
\newblock \href {https://doi.org/10.18653/v1/2022.emnlp-main.725} {Looking at the overlooked: An analysis on the word-overlap bias in natural language inference}.
\newblock In \emph{Proceedings of the 2022 Conference on Empirical Methods in Natural Language Processing}, pages 10605--10616, Abu Dhabi, United Arab Emirates. Association for Computational Linguistics.

\bibitem[{Seddik et~al.(2024)Seddik, Chen, Hayou, Youssef, and DEBBAH}]{seddik2024how}
Mohamed El~Amine Seddik, Suei-Wen Chen, Soufiane Hayou, Pierre Youssef, and Merouane~Abdelkader DEBBAH. 2024.
\newblock \href {https://openreview.net/forum?id=t3z6UlV09o} {How bad is training on synthetic data? a statistical analysis of language model collapse}.
\newblock In \emph{First Conference on Language Modeling}.

\bibitem[{Shumailov et~al.(2024)Shumailov, Shumaylov, Zhao, Papernot, Anderson, and Gal}]{shumailov2024ai}
Ilia Shumailov, Zakhar Shumaylov, Yiren Zhao, Nicolas Papernot, Ross Anderson, and Yarin Gal. 2024.
\newblock Ai models collapse when trained on recursively generated data.
\newblock \emph{Nature}, 631(8022):755--759.

\bibitem[{Singhal et~al.(2024)Singhal, Goyal, Xu, and Durrett}]{singhal2024a}
Prasann Singhal, Tanya Goyal, Jiacheng Xu, and Greg Durrett. 2024.
\newblock \href {https://openreview.net/forum?id=G8LaO1P0xv} {A long way to go: Investigating length correlations in {RLHF}}.
\newblock In \emph{First Conference on Language Modeling}.

\bibitem[{Susser and Seeman(2024)}]{Susser2024-SUSCPF}
Daniel Susser and Jeremy Seeman. 2024.
\newblock Critical provocations for synthetic data.
\newblock \emph{Surveillance and Society}, 22(4):453--459.

\bibitem[{Touvron et~al.(2023)Touvron, Martin, Stone, Albert, Almahairi, Babaei, Bashlykov, Batra, Bhargava, Bhosale et~al.}]{touvron2023llama}
Hugo Touvron, Louis Martin, Kevin Stone, Peter Albert, Amjad Almahairi, Yasmine Babaei, Nikolay Bashlykov, Soumya Batra, Prajjwal Bhargava, Shruti Bhosale, et~al. 2023.
\newblock Llama 2: Open foundation and fine-tuned chat models.
\newblock \emph{arXiv preprint arXiv:2307.09288}.

\bibitem[{Wang et~al.(2023)Wang, Kordi, Mishra, Liu, Smith, Khashabi, and Hajishirzi}]{wang-etal-2023-self-instruct}
Yizhong Wang, Yeganeh Kordi, Swaroop Mishra, Alisa Liu, Noah~A. Smith, Daniel Khashabi, and Hannaneh Hajishirzi. 2023.
\newblock \href {https://doi.org/10.18653/v1/2023.acl-long.754} {Self-instruct: Aligning language models with self-generated instructions}.
\newblock In \emph{Proceedings of the 61st Annual Meeting of the Association for Computational Linguistics (Volume 1: Long Papers)}, pages 13484--13508, Toronto, Canada. Association for Computational Linguistics.

\bibitem[{Wei et~al.(2024)Wei, Huang, Lu, Zhou, and Le}]{wei2024simplesyntheticdatareduces}
Jerry Wei, Da~Huang, Yifeng Lu, Denny Zhou, and Quoc~V. Le. 2024.
\newblock \href {https://arxiv.org/abs/2308.03958} {Simple synthetic data reduces sycophancy in large language models}.
\newblock \emph{Preprint}, arXiv:2308.03958.

\bibitem[{Whitney and Norman(2024)}]{whitney2024real}
Cedric~Deslandes Whitney and Justin Norman. 2024.
\newblock Real risks of fake data: Synthetic data, diversity-washing and consent circumvention.
\newblock In \emph{The 2024 ACM Conference on Fairness, Accountability, and Transparency}, pages 1733--1744.

\bibitem[{Williams et~al.(2018)Williams, Nangia, and Bowman}]{williams-etal-2018-broad}
Adina Williams, Nikita Nangia, and Samuel Bowman. 2018.
\newblock \href {https://doi.org/10.18653/v1/N18-1101} {A broad-coverage challenge corpus for sentence understanding through inference}.
\newblock In \emph{Proceedings of the 2018 Conference of the North {A}merican Chapter of the Association for Computational Linguistics: Human Language Technologies, Volume 1 (Long Papers)}, pages 1112--1122, New Orleans, Louisiana. Association for Computational Linguistics.

\bibitem[{Yang et~al.(2020)Yang, Malaviya, Fernandez, Swayamdipta, Le~Bras, Wang, Bhagavatula, Choi, and Downey}]{yang-etal-2020-generative}
Yiben Yang, Chaitanya Malaviya, Jared Fernandez, Swabha Swayamdipta, Ronan Le~Bras, Ji-Ping Wang, Chandra Bhagavatula, Yejin Choi, and Doug Downey. 2020.
\newblock \href {https://doi.org/10.18653/v1/2020.findings-emnlp.90} {Generative data augmentation for commonsense reasoning}.
\newblock In \emph{Findings of the Association for Computational Linguistics: EMNLP 2020}, pages 1008--1025, Online. Association for Computational Linguistics.

\end{thebibliography}

\end{document}